# scientific reports

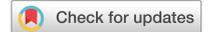

OPEN

# Predicting wind-driven spatial deposition through simulated color images using deep autoencoders

M. Giselle Fernández-Godino✉, Donald D. Lucas✉ & Qingkai Kong

For centuries, scientists have observed nature to understand the laws that govern the physical world. The traditional process of turning observations into physical understanding is slow. Imperfect models are constructed and tested to explain relationships in data. Powerful new algorithms can enable computers to learn physics by observing images and videos. Inspired by this idea, instead of training machine learning models using physical quantities, we used images, that is, pixel information. For this work, and as a proof of concept, the physics of interest are wind-driven spatial patterns. These phenomena include features in Aeolian dunes and volcanic ash deposition, wildfire smoke, and air pollution plumes. We use computer model simulations of spatial deposition patterns to approximate images from a hypothetical imaging device whose outputs are red, green, and blue (RGB) color images with channel values ranging from 0 to 255. In this paper, we explore deep convolutional neural network-based autoencoders to exploit relationships in wind-driven spatial patterns, which commonly occur in geosciences, and reduce their dimensionality. Reducing the data dimension size with an encoder enables training deep, fully connected neural network models linking geographic and meteorological scalar input quantities to the encoded space. Once this is achieved, full spatial patterns are reconstructed using the decoder. We demonstrate this approach on images of spatial deposition from a pollution source, where the encoder compresses the dimensionality to 0.02% of the original size, and the full predictive model performance on test data achieves a normalized root mean squared error of 8%, a figure of merit in space of 94% and a precision-recall area under the curve of 0.93.

Spatial patterns influenced by wind-driven dynamics are common in geosciences. Examples include algal blooms in the ocean surface[1], Aeolian sand dunes[2], and atmospheric dispersion plumes of air pollution[3], volcanic ash[4], and wildfire smoke[5], among others. Images and maps of spatial patterns collected from satellites and other remote sensing platforms encode the physical relationships between the prevailing wind conditions and resulting patterns[6]. Note that spatial pattern refers to an associated plume at a particular time or an integral over time after the source release. That is, there is no time variation involved. These images (snapshots) can be used to build data-driven models that predict new spatial patterns given wind inputs. A predictive model fully trained on images implicitly contains the relevant physical processes without using a mathematical model that relies on approximations and assumptions. Another advantage is the speedup in predictions compared with physics-based models, which discretize space into many cells and solve expensive mathematical equations. Figure 1 shows aerial pictures of some examples of these atmospheric plumes. Figure 1a shows a methane plume detected by the hyperspectral imaging spectrometer of NASA's Carbon Mapper satellite. The leak was later confirmed and repaired by the operator[7]. Figure 1b, c and d were taken by the Operational Land Imagers from satellites. These cameras measure in the visible, near-infrared, and short-wave infrared portions of the spectrum[8]. Spectrometers and thermal sensors are particularly useful when the quantity of interest is not visible to humans, as in the case of the methane plume in Fig. 1a. Cameras can be used to capture visible plumes. Satellites obtain images of these events daily; however, leveraging them to build predictive machine learning models has not been thoroughly exploited yet. This is our motivation, and our work is a starting point in this direction.

Effective dimensional reduction facilitates additional avenues for research and analysis of spatial patterns. Suppose most of the variability in a set of spatial patterns can be explained by a relatively small number of principal components. In that case, the dimensionality of the spatial problem can be greatly reduced. For the past six and a half decades[9], the meteorological community has relied upon principal component analysis (PCA) for analyzing and decomposing features in spatial patterns. For example, Higdon et al.[10] showed how computationally

Lawrence Livermore National Laboratory, 7000 East Ave, Livermore, CA 94550, USA. ✉email: fernandez48@llnl.gov; lucas26@llnl.gov

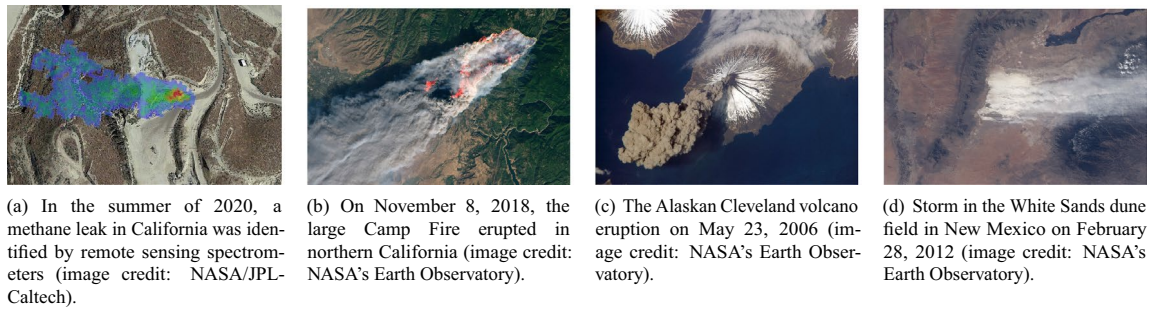

(a) In the summer of 2020, a methane leak in California was identified by remote sensing spectrometers (image credit: NASA/JPL-Caltech).

(b) On November 8, 2018, the large Camp Fire erupted in northern California (image credit: NASA's Earth Observatory).

(c) The Alaskan Cleveland volcano eruption on May 23, 2006 (image credit: NASA's Earth Observatory).

(d) Storm in the White Sands dune field in New Mexico on February 28, 2012 (image credit: NASA's Earth Observatory).

**Figure 1.** Multiple examples of atmospheric plumes detected by satellites using spectrometers, sensors, and imagers.

expensive Bayesian calibration and inference methods could be applied to images or data from complex physics models after first reducing the dimensionality using PCA. In Francom et al.[11], PCA is applied to spatiotemporal plume data before training a statistical model to emulate wind-driven atmospheric transport. Despite their success, a known issue associated with these orthogonal approximations is their inability to model non-linear problems. Recent advances in computational power and data availability have facilitated the path for the scientific community towards big data science and deep learning models[12]. An advantage of these models is the portability and speedup in predictions that can be achieved if compared with high-fidelity simulations or experiments, doing so with comparable accuracy.

The first and most straightforward artificial neural networks are multi-layer perceptrons (MLPs). These are comprised of layers with multiple neurons where the information travels only in the forward direction. They collectively learn from the inputs to predict the desired outputs in a process called weight optimization[13–15]. On the other hand, convolutional neural networks (CNNs) contain one or more convolutional layers that can be either connected or pooled[16]. While MLPs are processed only in a forward direction, CNNs take images as input data and refer to the same data multiple times when training. Due to their forward-facing nature, MLPs perform better for data classification tasks (text, sequences, and physical quantities) than for image analysis. CNNs take their name from the linear mathematical operation between matrices called convolution, which is essentially an averaging process. A CNN can have multiple flavors of layers, including convolutional layers, non-linearity layers, pooling layers, and concatenated fully-connected layers[17–19]. In recent years, CNNs have been widely used in computer vision problems due to their ability to learn hidden, non-linear features in data without a formal feature extraction procedure[20–23], making them a great candidate for image classification. These advantages make CNNs attractive for the analysis of patterns in geosciences. One example is the work of Sadeghi et al.[24], where a CNN-based model that uses two channels, water vapor and infrared quantitative precipitation estimation, to predict spatial precipitation patterns is developed.

CNNs are often used to build a powerful dimensional reduction and feature extraction technique called autoencoder[25]. A single-layered autoencoder with a linear activation function is very similar to a PCA[26], and therefore autoencoders can be thought of as a non-linear extension of PCA. Recent examples demonstrate how autoencoders are actively used as feature extractors in plasma physics[27] and seismology[28]. Autoencoders can model complex non-linear functions because their layers are composed of activation functions, such as sigmoid, ReLU, and tanh[29], which allows for non-linear predictions outperforming PCA for non-linear data. Another advantage of autoencoders, compared with linear transformations, is how flexible they are for model creation (number of layers, type of loss functions, kind of training algorithms, choice of the optimizer, among others), which can be constructed for different problems and needs (high-resolution, low-fidelity, large models, portable size models, among others).

The novelty of our work is the use of images and pixel information, instead of physical quantities, for predicting spatial patterns. In this paper, we show that an autoencoder consisting of multiple CNN layers with a relatively small number of network parameters ($\approx 50,000$) can be leveraged to predict the spatial patterns associated with plumes blowing in different directions and originating from different locations. In this work, we refer to (i) hyperparameters as non-physical quantities that can be specified while constructing the machine learning architecture (e.g., batch size, learning rate, number of neurons in each layer), to (ii) network parameters as non-physical quantities that are optimized during training and that cannot be controlled directly (e.g., weights value), and finally, to (iii) physical parameters as the specifics of the physics problem being solved (e.g., source location, wind direction, wind speed). In the Modified National Institute of Standards and Technology (MNIST) dataset of grayscale images (single channel) of handwritten digits[30], commonly used for benchmarking autoencoders, the digits are always located in the image center and have a roughly constant line width. Unlike the MNIST dataset, our color images (three channels) include spatial patterns that are not image-centered and where the object width changes based on the wind conditions (plumes are short and wide or long and narrow), adding an extra challenge. Another difference is that instead of ten objects (digits from zero to nine), we only have a unique object, a rotating and moving spatial pattern.

To our knowledge, an adequate set of real images for training CNNs that display wind-driven deposition patterns, or steady-state plumes, for different wind directions and release points does not exist. We, therefore, leverage deposition data from an existing set of thousands of computational fluid dynamics simulations described in the following two sections. We convert the deposition patterns into RGB images, which we assume were collected by a hypothetical imaging device suspended high above the surface. To connect our work to future applications, advances in optical metrology[31] and multispectral and hyperspectral[32,33] cameras offer the possibility of remotely imaging certain types of deposited materials. We firmly believe that our methodology can be extended and be predictive for real-world images of plumes or deposition with minor modifications. Our future goal is to develop image-based, data-driven models to predict the spatiotemporal evolution of real-world plumes that are directly observable with cameras, like the ones shown in Fig. 1b,c and d. In this work, we focus on the simpler problem of predicting only static spatial patterns (i.e., patterns that do not vary with time). If the plumes in Fig. 1 were subject to constant continuous emissions and steady winds, then steady-state patterns would emerge for each of these examples. Static spatial patterns can also arise for dispersion events with unsteady winds and short-duration emissions when the plume particles deposit by gravity to the surface and remain immobilized. Although deposition patterns like these are time-invariant, they are highly correlated with transient plumes because they represent a time-integrated history of where the overlying plume has traveled.

**The physics problem.** This paper's physics problem of interest is a two-dimensional, spatial pattern formed from a pollutant that has been released into the atmosphere and dispersed for up to an hour while undergoing deposition to the surface. The pollutant's release location $(s_x, s_y)$ is assumed to occur anywhere in a two-dimensional domain of 5000 m × 5000m. The release is initialized from a small bubble at time zero that is centered five meters above the surface, has a radius of five meters, and has internal momentum that causes it to expand radially and rise to a height of about 100 meters within the initial minute of simulation time. The same bubble source was used for all the simulations as a simplification. Only the $(s_x, s_y)$ coordinates of the locations of the bubble source are relevant. To capture variations in the height of the source across different simulations or images, we would incorporate $s_z$ as an extra scalar input alongside $(s_x, s_y)$. All the realizations used unit mass releases, and the resulting deposition patterns can be scaled proportionately for other mass amounts. The time scale of the simulated data represents the cumulative mass deposited on the surface for one hour. The pollutant is blown in a direction controlled by the large-scale atmospheric inflow winds expressed as a wind speed ($w_s$), which varies from 0.5 to 15 m/s, and wind direction ($w_d$), which can be anywhere in the interval [0, 360)° degrees following standard mathematical convention. For training purposes, however, we use $w_u = w_s \cos w_d$ and $w_v = w_s \sin w_d$, that is wind velocity components projected onto the $x$ and $y$ axes. We assume that the spatial patterns were collected by a hypothetical imaging device that records the magnitude of the logarithm of deposition as a red, green, and blue (RGB) color image with channels containing integer values ranging from 0 to 255. The goal is to predict a deposition image given its associated release location and wind velocity (four scalar quantities). In other words, we are interested in the following mapping: $[s_x, s_y, w_u, w_v] \rightarrow$ [height × width × RGB channel].

**The data.** The data is obtained from simulations and later post-processed to make it adequate for machine learning training. Given large-scale winds as an inflow boundary condition, the computational fluid dynamics (CFD) code Aeolus[34] uses hundreds of millions of grid cells to simulate fluid flow and material transport in complex, three-dimensional environments at high resolution, accounting for turbulence from structures, terrain features, and obstacles and predicting deposition on the ground and other surfaces. For demonstration purposes, megapixel deposition images were obtained by processing the output of Aeolus simulations, which were run using a resolution of $(x, y, z) = 1000 \times 1000 \times 100$ cells, each cell representing 5 m × 5 m × 5 m. Within Aeolus, pollutant concentration and deposition values are calculated by releasing and transporting Lagrangian particles of specified masses and sizes within the flow field. Particles that intersect the ground or other surfaces through turbulence or gravitational settling are removed from the atmosphere and recorded as deposition having units of mass per area. As previously noted, the releases were modeled as small, rising bubbles of mass that get carried by the winds about a minute into the simulations. Note that the actual deposition values are not considered in this paper.

The entire dataset, created by running Aeolus multiple times, contains 12,000 deposition images. The data images are stored as [number of images, height, width, RGB channels] = [12,000, 1000, 1000, 3]. Each megapixel image shows the spatial deposition pattern of a unique release scenario in Aeolus. The images are obtained by running an Aelous simulation with different release locations and inflow winds, $[s_x, s_y, w_u, w_v]$, using Latin hypercube sampling technique[35] within the design of experiment. Because we used the Aeolus CFD model as a source of data, we have enough images to train our model. However, for sparse data, data augmentation can leverage data symmetries[36] and improve performance. For predicting deposition patterns, the data can potentially be augmented for different wind directions by rotating the spatial plume pattern. In practice, this is not always possible due to the presence of terrain-based asymmetries in transport and dispersion. Because Aeolus accounts for terrain changes across the grid while solving the fluid equations, we do not have to include terrain or elevation explicitly in our data-driven model. However, we are investigating ways to embed terrain information as a boundary condition in CNNs[37] that may help improve predictions using real-world images. The Python *rainbow* colormap is used to create the RGB images for training and testing the autoencoder. Figure 2 shows the mapping between the RGB values, colormap, and deposition values. As previously noted, RGB pixel colors are associated with the logarithm of the deposition values.

**The deep-learning model.** The architecture of the deep learning model consists of three submodels: (i) the bottleneck model, (ii) the corrector model, and (iii) the autoencoder's decoder. This full prediction model composed of three parts is called the autoencoder-based model (ABM). Each submodel was used to perform the tasks it had been created and optimized to accomplish. The bottleneck model, an MLP approach, performs a link function between location, wind velocity, and reduced space obtained through the autoencoder's encoder. Although the bottleneck model does a fantastic job predicting the reduced space, it is not good enough for the decoder to recover the spatial pattern image with an accurate representation of high deposition areas (red-colored areas near the source), which is required in our application. The corrector or denoiser model can take the blurry predicted reduced space and improve it to serve the purpose at a meager cost. The combination of these three specialized models gave us better performance, efficiency, and lower memory print than the other tested model options, to name a few K-nearest neighbors (KNN)[38], single deep CNN-based model[39] and linear regression[40]. In other words, the autoencoder is used to reduce the spatial elements in the original megapixel image, allowing a successful training of a fully-connected bottleneck model linking geographic and meteorological scalar input quantities to the encoded space. Although accurate, the bottleneck model predictions are biased near the source (red color), which are corrected using the denoiser model. Once this is achieved, the decoder is used to recover an estimate of the desired megapixel image, completing the loop. Figure 3 shows a schematic of the ABM.

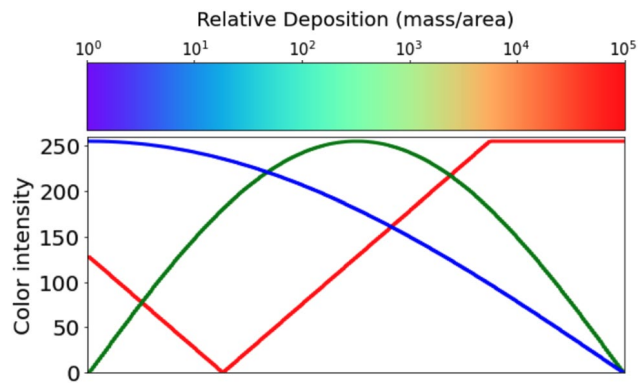

**Figure 2.** Red, green, and blue (RGB) values and their corresponding magnitude of deposition for the Python *rainbow* colormap used to plot the images used for training purposes.

Without an autoencoder, training a fully connected model with a single hidden layer and three million neurons to connect four scalars to a megapixel image would involve roughly a trillion network parameters and a petabyte of storage for training on batches of 128 images (note that even if we use a single image per batch, we still need tens of terabytes), which is intractable for today's computer power. The rule of thumb used for the previous calculation is that the neurons needed are $2/3\times$ input size + output size[41]. Furthermore, even in the hypothetical case that we can build this model, the prediction times associated would be impractical. Another possibility is to train a single model using one or more initial fully connected layers, followed by a reshaping layer that transforms the initial one-dimensional tensor into a two-dimensional tensor, and then one transposed convolutional layers (see Ahsan et al.[17]). We found out, however, that using multiple disconnected models instead gave us better performance while allowing for more flexibility because independent models can be improved separately without affecting the remaining architecture.

## Results

**Autoencoder vs. ABM reconstruction.** This section compares the autoencoder reconstruction and the overall ABM prediction performance. The performance of the autoencoder, which is, in actuality, the decoder performance (the encoder only reduces the dimension), can only be equal to or better than the performance of the ABM (bottleneck model + corrector + decoder) because the later includes, besides the decoder error,

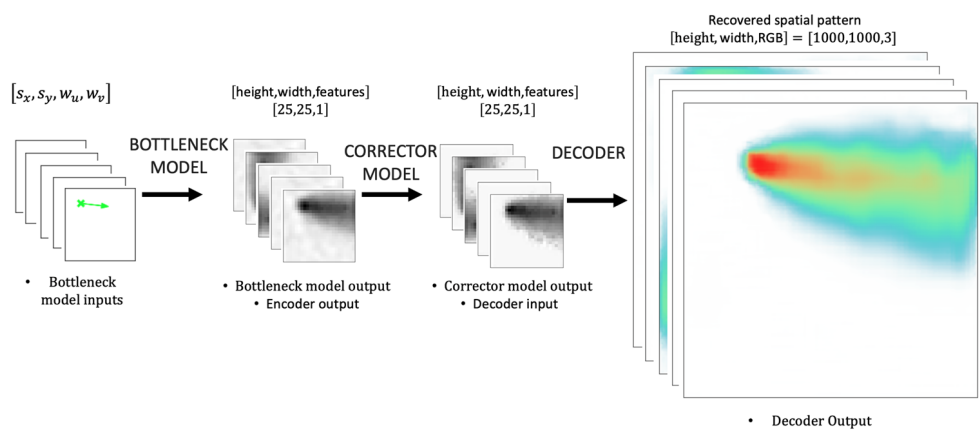

**Figure 3.** Schematic of the ABM encompassing the bottleneck model, corrector, and decoder.

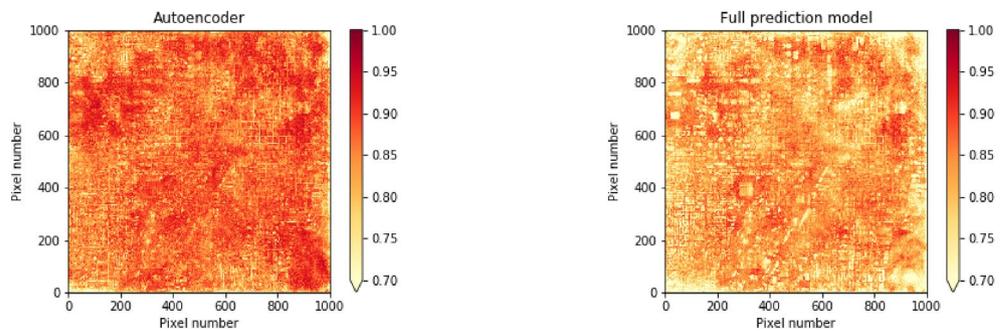

(a) Correlations between the true images and the decoder reconstruction (not ABM) for each pixel. The mean correlation value is 0.86. The minimum correlation value in a pixel is 0.18, and the maximum is 0.98.

(b) Correlations between the true images and ABM predictions (bottleneck + corrector + decoder) for each pixel. The mean correlation value is 0.82. The minimum correlation value in a pixel is 0.20, and the maximum is 0.95.

**Figure 4.** Mean pixel-level correlation values between the ground truth and predicted images for 300 test cases. The correlation shown per pixel is the average of the three channels per image.

the bottleneck model error plus the corrector error. Figure 4 shows the pixel-level Pearson correlation between truth and predicted images for 300 test cases. Note that the model has never seen test cases during training. Figure 4a shows the correlations between the autoencoder predictions and the corresponding ground truth images, while Fig. 4b shows the correlation between the ABM and the corresponding ground truth images. The figures show that the performance is good throughout the spatial domain, with almost all the pixels having correlations higher than 0.9, except for the smaller correlations near the lower-left and upper-right corners. The mean correlation value is larger than 0.8 in both figures. Observe that the correlations in Fig. 4b are lower than those in Fig. 4a as expected.

**Qualitative ABM performance assessment.** This section shows qualitative results and quantitatively assesses the model performance using two conceptually different metrics. Figure 5 shows, in each row, predictions for a randomly selected test case (six cases total). In the first column, the four scalar inputs are depicted by the green cross symbolizing the location of the source release and a wind vector arrow that shows the wind's speed (arrow length) and direction (arrow direction). The second column shows the latent space associated with each case predicted by the bottleneck model. The third column shows the corrected latent space. The fourth column shows the overall ABM predictions. Finally, the fifth column shows the ground truth images.

A comparison between the bottleneck model output (column two) and the corrector model output (column three) in Fig. 5 suggests that the corrector's main contribution is to removing noise in the white areas. A comparison between the last two columns in Fig. 5, ABM model prediction (fourth column) and ground truth (last column), shows outstanding performance.

**Quantitative ABM performance assessment.** In this section, the ABM performance was compared against two models. (i) The reference model, which is not a trained model, takes the source location and wind velocity as inputs and returns random patterns from the training data. This model can be interpreted as the diagonal line on a receiver operating characteristic (ROC)[42] curve. (ii) The baseline model, K-nearest neighbor (KNN)[38] approach using a single neighbor (K = 1). KNN approach identifies the closest input conditions between the desired input and the training inputs using the Euclidean distance. The KNN approach prediction is the spatial pattern associated with the closest input conditions within the training dataset.

The performance was assessed using two metrics. (i) The normalized root mean squared error (NRMSE), which considers all image pixels in the error calculation and is a metric commonly used for computer vision and image recognition applications because it gives an overall performance metric, and (ii) the figure of merit in space (FMS)[43], which is widely used in atmospheric sciences because it mainly focuses on how well the model predicts spatial patterns. FMS is also known as Jaccard index[44] and intersection over union criterion[45]. FMS only assesses how well non-white pixels (binary spatial pattern), between the true image and the predicted image, overlap, making it highly sensitive to rotations[46]. RMSE, which considers the magnitude values instead of binary ones, as FMS does, was included as a metric to have a more comprehensive measure of how the model performs throughout the domain. These two metrics complement each other giving a comprehensive performance score for the model.

Equation (1) shows how the NRMSE is calculated for a pair of predicted and true images (case $i$) where $\| \cdot \|_2$ denotes the $l_2$-norm. A perfect prediction would have an NRMSE of 0. Figure 6a shows the NRMSE histograms for the ABM, the baseline model, and the reference model. The histograms were built considering the 300 cases set aside for testing purposes. The figure includes the mean NRMSE for the three models: 8% RMSE for the ABM, 12% for the baseline model (KNN approach), and 22% for the reference model (random plume). NRMSE considers white pixels, that is, pixels where no deposition exists. The images in our dataset are approximately 85% white pixels; hence even if the prediction and the true pattern do not overlap, they still have around 70% of matching pixels explaining the mean NRMSE for the reference model (random plume) being only 22%.

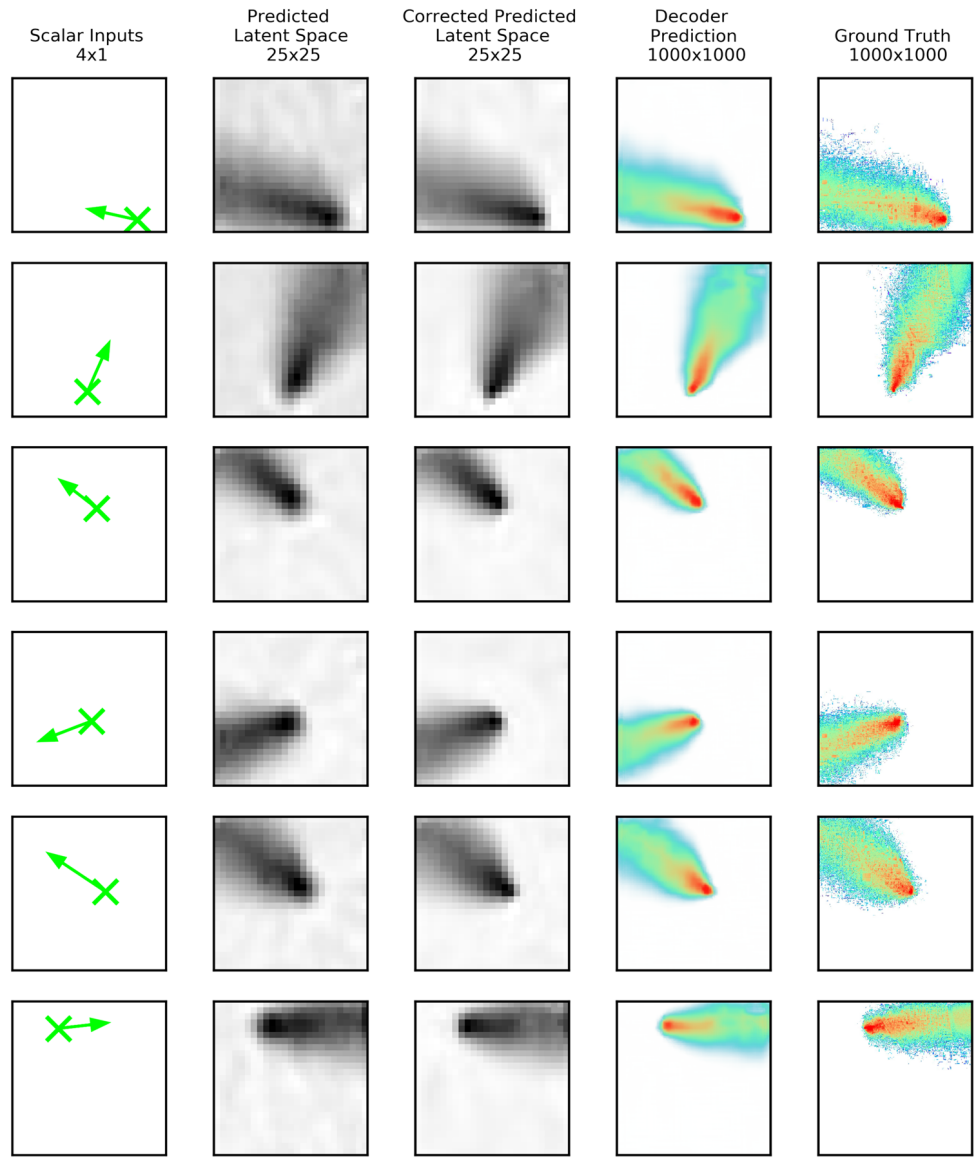

**Figure 5.** Images corresponding to different stages of the ABM prediction process for six randomly selected test cases shown in each row. The first column shows a schematic of the scalar inputs source location (cross), wind speed (arrow length), and wind direction (arrow orientation). The second column shows the predicted latent space using the bottleneck model. The third column shows the corrected latent space. The fourth column shows the final prediction using the decoder. Finally, the fifth column shows the ground truth. The colors of the pixels in the two rightmost columns denote the logarithm of deposition.

$$\text{NRMSE}_i = 100 \times \frac{\|\text{pixels(true image)}_i - \text{pixels(predicted image)}_i\|_2}{\|\text{pixels(true image)}_i\|_2} \quad (1)$$

Equation (2) shows how the FMS is calculated for a pair of images (case $i$) where ∩ and ∪ are the set notation of intersection and union, respectively. The FMS captures the fractional overlap of two spatial patterns. Perfect prediction results in an FMS of 100%. The *pattern* referred in Eq. (2) is created as follows: (1) convert color images into grayscale by taking the average of each of the image three pixels values, (2) set the values larger than 240 (white pixels) to "NaN" and (3) convert values smaller or equal to 240 (gray and black pixels) to zero (black). We selected 240 as the optimal threshold because it maximizes the performance of the ABM model predictions. Then, using Eq. (2), these converted images (spatial patterns) are used to compute the intersection divided by the union of the zero values (black pixels). Note that NaNs are ignored during this computation.

Figure 6b shows the FMS histograms for the ABM, the baseline, and reference models. The histograms were built considering the 300 cases set aside for testing purposes. The figure includes the mean FMS for the three models: 94% FMS for the ABM, 93% for the baseline model (KNN approach), and 73% for the reference model (random plume). The area under a ROC (ROC AUC)[47] is commonly used as a machine learning performance metric. When the data is imbalanced, however, the area under a precision-recall (PR) curve (PR AUC)[48] is

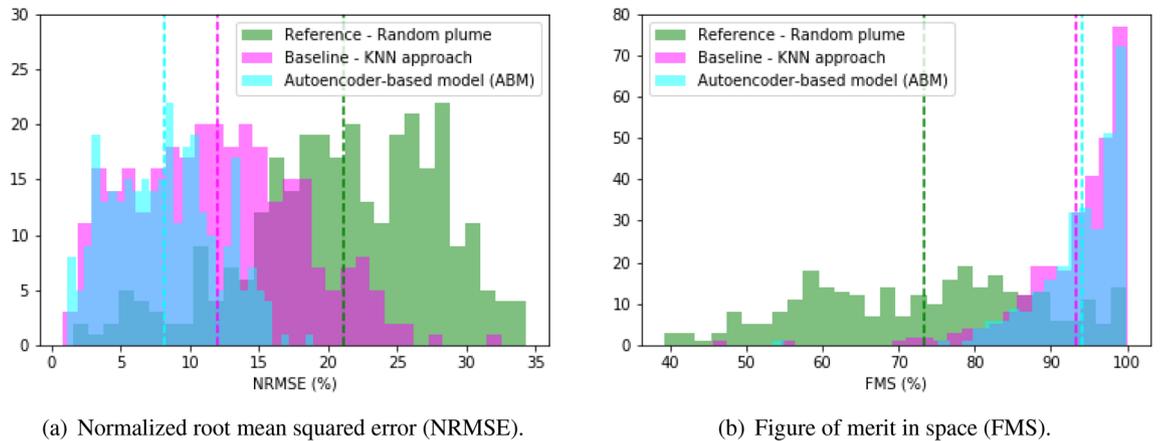

(a) Normalized root mean squared error (NRMSE).  (b) Figure of merit in space (FMS).

**Figure 6.** Performance metrics. The results are shown as histograms, highlighting the mean for each model, namely ABM, baseline, and reference. Each count corresponds to one of the 300 test cases.

preferred rather the ROC AUC because the former does not consider the performance on the negative class. On average, the images in our dataset have a 74% of white pixels (negative class) against an average of 26% of black pixels (positive class) making it an imbalanced dataset. The mean PR AUC for each of the three models is 0.93 for the ABM, 0.79 for the baseline model (KNN approach), and 0.14 for the reference model (random plume). The PR AUC curve is related to the FMS score in that the intersection component in Eq. (2) captures true positives, while the non-overlapping areas in the union capture false negatives and false positives.

$$\text{FMS}_i = 100 \times \frac{\text{true pattern}_i \cap \text{predicted pattern}_i}{\text{true pattern}_i \cup \text{predicted pattern}_i} \qquad (2)$$

Although a complete comparison of autoencoders with traditional orthogonal function decomposition is out of scope for this paper, we computed performance metrics for multiple PCA-based models to compare against the ABM presented in this work. The considered PCA variations taken into consideration in the analysis were PCA[49], Sparse PCA[50], and Truncated PCA[51]. Based on this analysis, we arrived at the following conclusions: For our dataset and models, (i) PCA has larger memory storage requirements than the decoder, 90 MB versus 0.211 MB for 32-bit floating point numbers. Hence PCA needs 426 times the storage space needed for the ABM. (ii) The PCA latent space (principal components) is harder to predict using a simple bottleneck model, possibly due to PCA's linear assumptions. (iii) Tensorflow-Keras makes it easier to train the ABM in batches and multiple GPUs, so very little memory in parallel is needed. Conversely, more memory is needed to train PCA. (iv) The FMS of the PCA-based reconstruction without a bottleneck model is an outstanding 99%. By comparison, the FMS for only the autoencoder is 97%. However, the NRMSE for PCA has a mean of 12% (versus 5% for the autoencoder). These results indicate that PCA performance compares with the autoencoder performance only on how well the predicted and ground truth patterns overlap, but the autoencoder performance is superior in predicting the value of pixel colors.

## Discussion

In this paper, we leveraged the spatial reduction that convolutional neural network-based autoencoders can achieve. We created an autoencoder ($\approx$ 50,000 network parameters) that reduces the dimension of a color image originally $1000 \times 1000$ pixels with red, green, and blue channels representing a two-dimensional deposition spatial pattern by 99.98%. This massive reduction enabled a fully-connected bottleneck model linking the source location and wind conditions to the reduced space achieving high accuracy. Then the decoder part of the autoencoder was used to recover the original resolution, completing the loop. The normalized root mean squared error of the model is 8%, the figure of merit in space is 94%, and the precision-recall area under the curve is 0.93. The performance metrics were computed on the test dataset (data never seen by the model during training). Ongoing and future work includes exploiting satellite and real-world images to train spatiotemporal predictive models of plumes in two dimensions[52] and three dimensions[53–55]. Future work also includes exploring other architectures, including a single model based on a combination of dense and convolutional layers. Although this work, as a proof-of-concept, is based on images of patterns obtained from physics-based models, future work will include using real-world images for training.

## Methods

**The autoencoder architecture.** We trained a CNN-based autoencoder (via Tensorflow-Keras). This autoencoder can be viewed as an identity operation with two parts: an encoder and a decoder. While the encoder takes the original image and reduces it to a latent space dimension, the decoder takes the reduced space vector and recovers the original image. In this case, the original RGB color image of size $1000 \times 1000$ pixels is reduced to a grayscale image of $25 \times 25$ pixels via the encoder. Note that the reduced image size is less than 0.02% of the original size. The original images can be recovered using the decoder. We tried different latent space sizes and obtained the best compromise between compression and performance at $25 \times 25$ pixels. If the latent space is too large, the bottleneck model cannot accurately link the four scalars to the latent space. Therefore, the bottleneck prediction of the reduced space is not good enough to allow the decoder an accurate recovery of the original image. On the other hand, if the size of the latent space is too small (too compressed), a fast-performing autoencoder cannot recover the original image.

The autoencoder can be divided into an encoder and a decoder, which can be used separately. In this case, we used the encoder as a compression tool where the input is a [1000, 1000, 3] image. The encoder output is a single feature map of size [25, 25, 1] that resembles the original color image in grayscale and with a much-reduced dimension (see "Supplementary information" for a brief discussion on the interpretability of the latent space). The ABM then leverages the decoder to recover the high-resolution color images. The autoencoder architecture was inspired by the Keras creator's tutorial[56], and the hyperparameters varied to find fast, accurate predictions using a reasonable memory footprint. For the gradient-based optimization, we used the Adam algorithm[57] along with an adaptive learning rate, set initially to 0.001 (initial default value). We used binary crossentropy as a loss function. In the studied case, being a regression problem, it would have been more natural to use the mean squared error (MSE) as a loss function. Although MSE loss resulted in a more accurate autoencoder with a more sophisticated latent space, it was not used here because we are looking for the best overall accuracy (bottleneck model + corrector model + autoencoder). We found that this more complex latent space, obtained using MSE, was not well represented by the bottleneck model, leading to poorer results.

The bottleneck model is an independent MLP (trained via scikit-lean) that maps the scalar input vector ([4, 1] ) to the reduced space ([25, 25, 1] ), and it is described in the next section. Figure 7 shows the autoencoder architecture schematically. The pixel values of the three color channels were normalized to the interval [0,1] before training. The data was divided into 10,530 training cases (used for training), 1170 validation cases (used to evaluate the loss at the end of each epoch), and 300 test cases (data never seen by the model). The model was trained using single precision data. The autoencoder model has a total of 52,634 network parameters, and it takes 0.36 s for a single prediction. Most of the details stated above are summarized in Table 1.

*The encoder.* The encoder comprises six two-dimensional convolutional layers and five two-dimensional max-pooling layers interspersed among the convolutional layers (11 layers total). All layers have padding with zeros evenly. All the convolutional layers have a rectified linear unit (ReLU) as the activation function. The first convolutional layer of the encoder has 32 kernels with a height and width of seven. The second has 16 kernels with a height and width of five. The third has eight kernels with a height and width of three. The fourth has four kernels with a height and width of three. The fifth has three kernels with a height and width of three. Finally, the encoder's sixth and last convolutional layer has one kernel with a height and width of three. The first, second, and third max-pooling layers have a pool size of two, the fourth has a pool size of five, and finally, the fifth layer has a pool size of one. The output of the encoder has dimensions of [25, 25, 1]. Observe that this is a factor of 4800 reduction. The encoder model has a total of 19,143 network parameters, takes 0.15 s for a single prediction, and reduces the original image to 0.02% of its size. The above details can be found in Table 1.

*The decoder.* The decoder is built using seven two-dimensional convolutional layers and six two-dimensional upsampling layers interspersed among the convolutional layers (13 layers total). All layers have padding with zeros evenly. All the convolutional layers have a ReLU as the activation function except the last layer, whose activation function is a sigmoid function. The first convolutional layer of the decoder has one kernel with a height and width of three. The second has three kernels with a height and width of three. The third has four kernels with a height and width of three. The fourth has eight kernels with a height and width of three. The fifth has 16 kernels with a height and width of five. The sixth has 32 kernels with a height and width of seven. Finally, the decoder's seventh and last convolutional layer has three kernels with a height and width of seven. The decoder model has a total of 33,491 network parameters, and it takes 0.21 s for a single prediction.

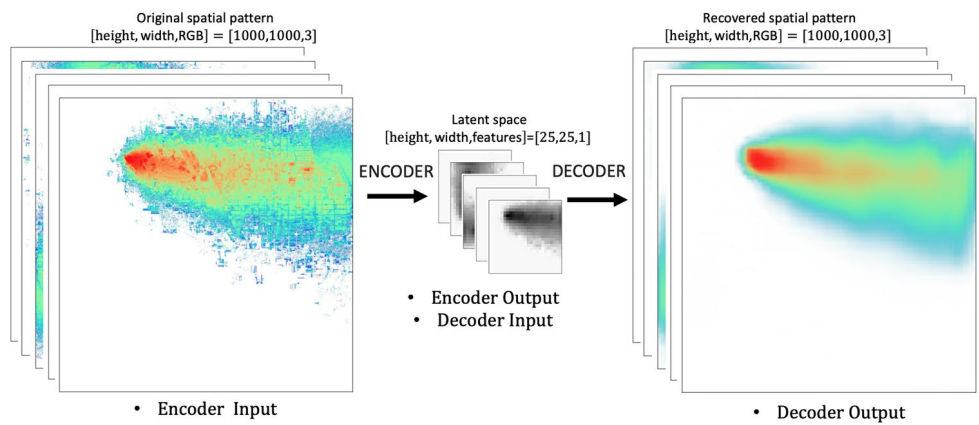

**Figure 7.** Schematic of the autoencoder architecture. The encoder receives the original image (shape [1000, 1000, 3]), and it reduces its size to 0.02% (shape [25, 25, 1]). The decoder can recover the original image from the reduced space quite accurately.

| Model | Submodel | Layer number | Layer type | Number of kernels | Kernel height × width | Activation function |
|---|---|---|---|---|---|---|
| Autoencoder | Encoder | 1 | Input | – | – | – |
| | | 2, 4, 6, 8, 10 | Convolutional 2D | 32, 16, 8, 4, 3 | 7 × 7, 5 × 5, 3 × 3, 3 × 3, 3 × 3 | ReLU |
| | | 3, 5, 7, 9, 11 | UpSampling 2D | – | – | – |
| | | 12 | Output/ Convolutional 2D | 1 | 3 × 3 | ReLU |
| | Decoder | 1 | Input | – | – | – |
| | | 2, 4, 6, 8, 10, 12 | Convolutional 2D | 1, 3, 4, 8, 16, 32 | 3 × 3, 3 × 3, 3 × 3, 3 × 3, 5 × 5, 7 × 7 | ReLU |
| | | 3, 5, 7, 9, 11, 13 | UpSampling 2D | – | – | – |
| | | 14 | Output/ Convolutional 2D | 3 | 7 × 7 | Sigmoid |
| Corrector | Encoder | 1 | Input | – | – | – |
| | | 2,3 | Convolutional 2D | 1, 64 | 7 × 7, 7 × 7 | ReLU |
| | | 4 | Output/Convolutional 2D | 128 | 7 × 7 | ReLU |
| | Decoder | 1 | Input | – | – | – |
| | | 2, 3 | Convolutional 2D | 64, 32 | 7 × 7, 7 × 7 | ReLU |
| | | 4 | Output/Convolutional 2D | 1 | 7 × 7 | ReLU |

**Table 1.** Architecture description of the models built using TensorFlow-Keras, the autoencoder and corrector models.

The first upsampling layer has an up-sampling factor of one. The second upsampling layer has an up-sampling factor of five. The third, fourth, and fifth upsampling layers have an up-sampling factor of two, and finally, the sixth upsampling layer has a pool size of one. The output of the decoder has dimensions of [1000, 1000, 3]. The sigmoid function used in the decoder's last layer forces the autoencoder to produce output values between 0 and 1 which is desired because we normalized the three-channel values to the interval [0,1]. The original pixel values can be recovered by multiplying by 255. The details described above are included in Table 1.

**The bottleneck architecture.** Once the decoder is built, the goal is to link the geographic and meteorological scalar input quantities to the reduced encoded space through the bottleneck model. The second step is to correct the bias using the corrector model. The motivation and details for both models are described in the following subsections.

*The bottleneck model.* To build the bottleneck model, we tested various model options, MLPs, linear regression, random forest, and XGboost. We found that MLPs outperform the other models' prediction accuracy while maintaining the desired prediction time. CNN-based models work great on images, which is why they are the best option for dimensional reduction, but our ABM inputs are a few scalars instead. MLPs have fully-connected layers and feed-forward architecture, which make them ideal for text or physical data (such as wind velocity and source location). The bottleneck dimension, 25 × 25 × 1, is now small enough to train a simple MLP model linking the wind conditions and release location (four scalar quantities) with the reduced space (see "Supplementary information" for a brief discussion on the interpretability of the reduced space). Note that training a fully connected model with a single hidden layer and three million neurons to connect four scalars to a megapixel color image (instead of the reduced 25 × 25 × 1 space) would involve an intractable trillion network parameters and a petabyte of storage and a prohibitive prediction time. The reduced space is fed flattened into the MLP (25 × 25 = 625) because we want a vector, not an image, in this case. For this task, we have used the neural network MLPRegressor model from the Python package scikit-learn version 0.24.2. The optimizer is Adam[57], and the activation is a sigmoid function. The sigmoid function performs better than other activation functions available in the package, such as ReLU or tanh, because the output data fed to the bottleneck model for training is scaled between 0 and 1; hence, by sigmoid (also known as the logistic function), we enforced the output to that range. We used default options except for the adaptive learning rate, which was initially set to 0.001 (initial default value), the L2 penalty alpha set to $10^{-5}$, three hidden layers with sizes of (300, 400, 300), batch size equal to 500, and the number of iterations without change set to 100. The training input data: source location in the horizontal direction, source location in the vertical direction, wind velocity in the horizontal direction, and wind velocity in the vertical direction ($s_x, s_y, w_u, w_v$) were standardized before training. The average bottleneck model $R^2$ score is 95% on the 300 test cases (data never seen before by the model).

*The corrector model.* Even if the bottleneck model described in the previous subsection had a great overall performance, it did not perform as desired in high deposition areas (red areas), hence the need to improve the latent space prediction with a corrector. The corrector model is a second CNN-based autoencoder model (via Tensorflow-Keras), inspired by denoising or imputing autoencoders[58,59]. Instead of reducing dimensions, as the autoencoder used to build the ABM, this autoencoder is used as a "denoiser" with a larger bottleneck dimension (also known as an overcomplete autoencoder).

The corrector model takes as input the bottleneck model output (shape [25, 25, 1]) and returns a corrected image (shape [25, 25, 1]). The corrector bottleneck dimension is $25 \times 25 \times 128$ (128 times larger than the input image dimension). This latent space extracts 128 features from the original image and recovers a corrected image. The corrector model was trained using the bottleneck model outputs (noisy data) as the input training data and the encoder outputs (original reduced space used as ground truth) as output training data. The pixel values were normalized to the interval [0,1] before training. The data was divided into 10,530 training cases (used for training), 1170 validation cases (used to evaluate the loss at the end of each epoch), and 300 test cases (data never seen by the model).

The corrector's encoder is built using two two-dimensional convolutional layers. All layers have padding with zeros evenly. All the convolutional layers have a ReLU as the activation function. The first convolutional layer of the corrector's encoder has 64 kernels with a height and width of seven. The second has 128 kernels with a height and width of seven. The output of the corrector's encoder has a shape of [25, 25, 128]. The corrector's decoder is built using three two-dimensional convolutional layers. All layers have padding with zeros evenly. All the convolutional layers have a ReLU as the activation function. The first convolutional layer of the corrector's decoder has 64 kernels with a height and width of seven. The second has 32 kernels with a height and width of seven. Finally, the third has a kernel with a height and width of seven. The output of the corrector's decoder has a shape of [25, 25, 1].

Most of the above details are summarized in Table 1. The original pixel values can be recovered by multiplying by 255. For the first-order gradient-based optimization, we used the Adam algorithm[57] and the mean squared error as a loss function. The corrector model has a total of 908,161 network parameters. Despite that the number of network parameters involved in the corrector is larger than in the main autoencoder, the computation time is negligible because it is applied to the reduced space (625 pixels instead $3 \times 6$ pixels). The average bottleneck model $R^2$ score is 93% on the 300 test cases (data never seen before by the model). Even if the corrector model $R^2$ score is slightly poorer (if compared with the $R^2 = 95\%$ resulting from using only the bottleneck model without the corrector discussed in the previous subsection), the overall accuracy of the full prediction model was improved, resulting in a mean NRMSE decrease of 2% and a mean FMS increase of 5%.

**The ABM.** The ABM model consists of aligning the three different independent models described in the previous sections. First, the bottleneck model connecting four scalar quantities ($s_x, s_y, w_u, w_v$) to the grayscale latent space of shape [25, 25, 1]; second, correcting the grayscale latent space of shape [25, 25, 1] through the corrector model obtaining a corrected grayscale latent space of shape [25, 25, 1]; and third connecting the estimated corrected latent space to the high-resolution prediction of the deposition spatial pattern color image of shape [1000, 1000, 3], through the decoder. Figure 3 shows a schematic representation of the process described. It is worth mentioning that the same split of training, validation, and testing cases was used throughout the training, validation, and testing process for each of the models (bottleneck, corrector, and autoencoder).

## Data availability
The data supporting this study's findings are available from the corresponding authors upon request.

## Code availability
The deep learning architecture, the trained model, and the statistical analysis are based on already published work and have been detailed in the document for the reader's reproduction. The Tensorflow version used was 2.5.0, and the Keras version was 2.2.4.

### Acknowledgements
This work was performed under the auspices of the United States Department of Energy (DOE) by Lawrence Livermore National Laboratory under contract DE-AC52-07NA27344 and public release number LLNL-JRNL-831189. This research was funded by the National Nuclear Security Administration, Defense Nuclear Nonproliferation Research and Development (NNSA DNN R &D). The authors acknowledge important inter-disciplinary collaboration with scientists and engineers from LANL, LLNL, MSTS, PNNL, and SNL. The authors thank Gemma J. Anderson, Akshay Gowardhan, and Nipun Gunawardena for their valuable help and guidance, and Lee G. Glascoe and Stephen C. Myers for supporting this publication.

### Author contributions
M.G.F.G.: Writing—original draft, Visualization, Formal analysis, Investigation, Conceptualization. D.D.L: Supervision, Project administration, Conceptualization, Methodology, Writing—review & editing. Q.K.: Supervision, Methodology, Writing—review & editing.

### Competing interests
The authors declare no competing interests.

### Additional information
**Supplementary Information** The online version contains supplementary material available at https://doi.org/10.1038/s41598-023-28590-4.

**Correspondence** and requests for materials should be addressed to M.G.F.-G. or D.D.L.

**Reprints and permissions information** is available at www.nature.com/reprints.

**Publisher's note** Springer Nature remains neutral with regard to jurisdictional claims in published maps and institutional affiliations.